\documentclass[11pt]{article}

\newif\ifarxiv
\arxivtrue

\usepackage[mathscr]{eucal}
\usepackage{epsfig,epsf,psfrag}
\usepackage{amssymb,amsfonts,latexsym}
\usepackage{amsmath}
\usepackage{subfigure}
\usepackage{graphicx}
\usepackage{ulem}

\usepackage{epstopdf}
\usepackage{bm,xcolor,url}
\usepackage{array}
\usepackage{verbatim}
\usepackage{algpseudocode, cite}
\usepackage{algorithm}
\usepackage{verbatim}
\usepackage{textcomp}
\usepackage{mathrsfs}
\usepackage[referable]{threeparttablex}
\usepackage{amsthm}
\usepackage{enumerate}
\usepackage{footnote}
\usepackage{enumitem}
\usepackage{xr}
\usepackage{mathtools}
\usepackage{stmaryrd}
\usepackage{wrapfig}
\usepackage{pdfpages}

\catcode`~=11 \def\UrlSpecials{\do\~{\kern -.15em\lower .7ex\hbox{~}\kern .04em}} \catcode`~=13 

\allowdisplaybreaks[3]

\newcommand{\tnorm}[1]{{\left\vert\kern-0.25ex\left\vert\kern-0.25ex\left\vert #1 
    \right\vert\kern-0.25ex\right\vert\kern-0.25ex\right\vert}}
\newcommand{\tnormt}[1]{{\vert\kern-0.25ex\vert\kern-0.25ex\vert #1 
    \vert\kern-0.25ex\vert\kern-0.25ex\vert}}

\newcommand{\defeq}{\triangleq}

\newcommand{\cM}{\mathcal{M}}

\newcommand{\calN}{\mathcal{N}}

\newcommand{\calX}{\mathcal{X}}

\newcommand{\bE}{\mathbf{E}}

\DeclareMathAlphabet{\mathbsf}{OT1}{cmss}{bx}{n}

\DeclareMathOperator*{\argmax}{arg\,max}
\DeclareMathOperator*{\argmin}{arg\,min}

\newcommand{\bone}{\mathbf{1}}

\DeclarePairedDelimiterX{\infdivx}[2]{(}{)}{%
  #1\;\delimsize\|\;#2%
}

\newcommand{\DKL}{D_{\rm KL}\infdivx}

\newtheorem*{theorem*}{Theorem}

\newtheorem*{assump*}{Assumption}

\theoremstyle{definition}

\theoremstyle{remark}

\newcommand{\qednew}{\nobreak \ifvmode \relax \else
      \ifdim\lastskip<1.5em \hskip-\lastskip
      \hskip1.5em plus0em minus0.5em \fi \nobreak
      \vrule height0.75em width0.5em depth0.25em\fi}

\newcommand{\mtrue}{m_\mathrm{true}}

\usepackage[colorlinks,linkcolor=blue,citecolor=blue,urlcolor=blue]{hyperref}
\usepackage[left=1in,right=1in,top=1in,bottom=1in,footskip=.25in]{geometry}

\usepackage{authblk}

\begin{document}

\ifarxiv
    \title{Bayesian Experimental Design for Symbolic Discovery}
    
    \author[a]{Kenneth L. Clarkson}
    \author[a]{Cristina Cornelio\footnote{Current address: Samsung AI Research, Cambridge, UK.}}
    \author[a]{Sanjeeb Dash}
    \author[a]{{Joao Goncalves}}
    \author[a]{Lior Horesh}
    \author[a]{Nimrod Megiddo}
    
    \affil[a]{IBM Research}
    \date{}

\fi 

\maketitle

\begin{abstract}
        This study concerns the formulation and application of Bayesian optimal experimental design to symbolic discovery, which is the inference from observational data of predictive models taking general functional forms.
        We apply constrained first-order methods to optimize an appropriate selection criterion, using Hamiltonian Monte Carlo to sample from the prior.
        A step for computing the predictive distribution, involving convolution, is computed via either numerical integration, or via fast transform methods.
\end{abstract}


\section{Motivation}

\ifarxiv
    This study concerns work done on Bayesian Optimal Experimental Design (OED) in the context of symbolic discovery\cite{cornelio2021ai}.
    The latter, also called symbolic regression, seeks to explain observational data using a model chosen from a range of functional forms, that may include general algebraic expressions, trigonometric functions, and so on.
    The model is typically chosen to balance criteria involving fidelity to the data, the simplicity of the model by some measure, and prior knowledge.
\fi 

The goal of optimal experimental design (OED) is to find the optimal design of a data acquisition system, so that the uncertainty in the inferred parameters, or some predicted quantity derived from them, is minimized with respect to a statistical criterion.
In the context of model discovery, a large body of work addresses the question of experimental design for pre-determined functional forms, and another body of research addresses the selection of a model (functional form) out of a set of
candidates.
In the context of symbolic regression, our aim is to devise an experimental setup that attends to both the functional form and the continuous set of parameters that defines the model behavior.
In realistic settings, experimental data may be restricted or costly, providing limited support for any given  hypothesis as to the underlying functional form.
Here we formulate a Bayesian framework for experimental design where joint model selection and parameter estimation are pursued.
Within that formulation, we show relationships, sometimes equivalence, of a few known selection criteria.
Following that we perform a preliminary validation study.
A key computational challenge will be computation of derivatives, for which we have used symbolic differentiation.
We also aim to explore both Markov Chain Monte Carlo (MCMC) as well more efficient sampling strategies as Hamiltonian Monte Carlo (HMC).

\section{Background art}

The topic of experimental design, and in particular design in a Bayesian framework, is rich and well-studied; an overview is given by \cite{ryan2016review}.
We will develop the framework appropriate to our setting and implementation, and return in \S\ref{subsec B further} to alternative choices and further work.

\section{Formulation}

\paragraph{Models, priors, and updates}
Denote the design (input) space as $\calX$. 
We have a set of models $\cM$, where each model $m\in\cM$ has parameters
$\theta_m\in\mathbb{R}^n$ for some $n$.
We assume that there is some ground-truth model $\mtrue$ with associated parameters $\theta_\mathrm{true}$, and the data we receive is 
\begin{equation}\label{eq resp meas}
    y(x) \defeq \mtrue(x,\theta_\mathrm{true}) + \epsilon,
\end{equation}
where $\epsilon\sim\calN(0,\sigma^2)$, for known variance $\sigma^2$.
Letting $\phi(z;\mu,\sigma^2)$ denote the density function
of a normal distribution $\calN(\mu,\sigma^2)$,
and writing $y(x)$ as $y$ where this is clear,
we can also write this as $p(y|m, \theta_m, x) = \phi(y; m(x, \theta_m),\sigma^2)$.

We consider $m\in\cM$ to be random, with prior $p(m)$,
and $\theta_m$ to be random, with prior $p(\theta_m)$.
These distributions express our belief that a given $m$ is $\mtrue$ (in functional form),
and our beliefs about the locations of the associated parameters.

An experiment here is a choice of $x$; given that choice, we learn
$y(x)$ as in \eqref{eq resp meas} above.
For that given $x$ and $y$, we can update $p(m)$ and $p(\theta_m)$
using Bayes rule, with posterior probability
\begin{align}\label{eq m post}
\tilde p(m | x, y) & = p(m) \frac{p(y|m,x)}{p(y|x)}, \mathrm{where}\nonumber \\
p(y|m,x) & = \bE_{\theta_m} p(y | m, \theta_m, x) = \bE_{\theta_m}\phi(y; m(x, \theta_m),\sigma^2), \mathrm{and} \\
p(y|x) & = \bE_m p(y|m,x) = \sum_{m\in\cM} p(m)p(y | m,x). \nonumber
\end{align}
Similarly,
\begin{equation}\label{eq theta post}
    \tilde p(\theta_m) = p(\theta_m)\frac{p(y | m, \theta_m, x)}{p(y | x)} = p(\theta_m)\frac{\phi(y; m(x, \theta_m),\sigma^2)}{p(y | x)}
\end{equation}

We will use Monte Carlo methods to sample from $p(\theta_m)$, for each $m$,
obtaining $S_m\subset\mathbb{R}^n$ for each $m\in\cM$, such that
(ideally) each member of $S_m$ has distribution $p(\theta_m)$.
For functions $f(\theta)$, we then estimate
\begin{equation}\label{eq E sum}
    \bE_{\theta_m} f(\theta_m)\approx \frac{1}{|S_m|}\sum_{\theta\in S_m} f(\theta).
\end{equation}

\paragraph{Selection criteria.}
We are looking for a design point $x\in\calX$ that is ``most informative,'' in some sense,
about the model. Picking a point $x^*$, we then update the posterior distributions
as in \eqref{eq m post} and \eqref{eq theta post}, and repeat.
That is, we make the resulting posterior into the prior for the next choice
of design point, so the \emph{current} $p(m)$ and $p(\theta_m)$ are based on a sequence
of choices $x^1,x^2,\ldots$, and corresponding $y^1,y^2,\ldots$, each
$y^t$ from \eqref{eq resp meas} and using the corresponding $x^t$.

\emph{Maximizing mutual information.}
A natural version of this general goal is to find $x$ that maximizes the mutual information
$I(y;m|x)$ between the response $y$ and the model $m$, conditioned on $x$. This approach, to find
\[
x^*_{\rm MI} \defeq \argmax_{x} I(y; m | x),
\]
was proposed in this form in, for example,~\cite{drovandi2014sequential}.

\emph{Minimizing model entropy.}
We could also consider selecting a point $x$ that minimizes
the entropy $H(m|y,x)$; however, since
$I(y; m | x) = H(m|x) - H(m| y,x) = H(m) - H(m|y,x)$,
we have
\[
x^*_{\rm ME} \defeq \argmin_x H(m|y,x) = x^*_{\rm MI}.
\]

\emph{Maximizing Jensen-Shannon divergence.}
A criterion based on the multi-way Jensen-Shannon divergence was proposed by
\cite{vanlier2014optimal}.
The form used was
\begin{align}\label{eq:JSD}
D_{\rm JS}(x) \defeq \bE_m[\DKL{(p(y|m,x)}{p(y|x)}],
\end{align}
where $\DKL{}{}$ is the Kullback-Liebler divergence.
That is, we seek to find the design point of
\begin{equation}\label{eq JS}
x^*_{\rm JS} \defeq \argmax_x D_{\rm JS}(x),
\end{equation}
that maximizes the expected (w.r.t. $m$)
divergence of $p(y|m,x)$ from the expectations (w.r.t. $m$) of those divergences.

However, as is well-known~\cite{wiki-MI, shulkind2018experimental}, the mutual information between
random variables $W$ and $Z$ satisfies
$I(W;Z) = \bE_W[\DKL{p(Z|W)}{p(Z)}]$,
so that taking $W=m$ and $Z=y$, and
conditioning everywhere on $x$ (and noting that $m$ is independent of fixed $x$),
the conditional mutual information
$I(y;m | x) = \bE_{m} [\DKL{(p(y|m,x)}{p(y|x)}]$, 
so we have also $x^*_{\rm JS} = x^*_{\rm MI}$.

\emph{Maximizing response entropy.}
Suppose the uncertainty of observation $y$ at a given point $x$ under model $m$
is independent of $m$ and $x$, that is, for any given $m,x, m', x'$,
$H(y|m,x) = H(y| m',x')$. This holds under our assumption of i.i.d. measurement
noise, since for all $m,x$, we have
$H(y|m,x) = H(\epsilon)$,
where $\epsilon$ is the error as in \eqref{eq resp meas} above.
Then, since also $I(y; m | x) = H(y|x) - H(y | m,x)$, and $H(y|m,x)$ is fixed
with respect to $x$, we have
\begin{equation}\label{eq resp ent}
x^*_{\rm RE}\defeq \argmin_x H(y|x) = \argmin_x H(y|x) - H(y | m,x) = \argmin_x I(y; m | x) = x^*_{\rm MI},
\end{equation}
a formulation apparently going back to~\cite{borth1975total}.
That is, the best $x$ makes the variation in the response distribution
$p(y|x)$ as large as possible.

\emph{Maximzing $\log\det$.}
So far, all selection criteria yield the same optimum
$x^*_{\rm MI} = x^*_{ME}=x^*_{\rm JS} = x^*_{RE}$; the simplest of these to compute seems
to be $x^*_{\rm RE}$, from \eqref{eq resp ent}.

A different idea that we have also explored is
again based on maximizing the dispersion of the response. Here we 
(conceptually) build a matrix $D(x)$ whose rows and columns are indexed by the set
$\{(m, \theta) \mid m\in\cM, \theta \in S_m\}$,
recalling that the $S_m$ are samples of $p(\theta_m)$, as used in \eqref{eq E sum}.
That is, this matrix $D(x)$  has $\sum_{m\in\cM} |S_m|$ rows and columns.
The entry of $D(x)$ for $(m, \theta)$ and $(m',\theta')$
is, using standard facts about the KL-divergence,
\begin{align*}
    & \DKL{ p(y|m,\theta,x)}{p(y|m',\theta',x)}
        \\ & = \DKL{\calN(m(x,\theta),\sigma^2)}{\calN(m'(x,\theta'),\sigma^2)}
        \\ & = \frac{\left(m(x;\theta),\sigma^2) - m'(x;\theta'),\sigma^2)\right)^2}{2\sigma^2},
\end{align*}
and the optimum is
\begin{equation}\label{eq logdet}
x^*_{\rm LD} \defeq \argmin_x -\log\det D(x).
\end{equation}

\section{Implementation considerations}

We implemented algorithms for 
finding $x^*_{\rm JS}$, which is the optimum for the design selection criterion based on
Jensen-Shannon \eqref{eq JS}; the equivalent one $x^*_{\rm RE}$ from \eqref{eq resp ent} based
on response entropy; the ``logdet'' optimum $x^*_{\rm LD}$ from \eqref{eq logdet};
and a variant of the logdet criterion. Below we will focus on the response entropy,
which seems the most promising.

\paragraph{Estimating $p(y|x)$.}
For computing $x^*_{\rm RE}$, we need an estimate of
\begin{equation}\label{eq def H}
    H(y|x) = -\int_y p(y|x)\log p(y|x) = -\bE_{y|x}[\log p(y|x)],
\end{equation}
using the convention (from the limit) that $0\log 0 = 0$.
Our estimate of $p(y|x)$ is, following \eqref{eq m post} and \eqref{eq E sum},
\begin{align}
p(y|x) & = \bE_m p(y|m,x) = \bE_m \bE_{\theta_m} \phi(y; m(x,\theta_m),\sigma^2)\nonumber
    \\ & \approx \frac{1}{|\cM|} \sum_{m\in \cM} p(m) \frac{1}{|S_m|}\sum_{\theta \in S_m} \phi(y; m(x,\theta_m),\sigma^2).\label{eq resp prob}
\end{align}
We implemented two ways to estimate $H(y|x)$: either
numerical integration of \eqref{eq def H} using \eqref{eq resp prob},
or via convolution, noting that
\[
p(y|m,x) \approx \frac{1}{|S_m|}\sum_{\theta \in S_m} \phi(y; m(x,\theta_m),\sigma^2)
    = \frac{1}{|S_m|} \bone_{S_m} * \phi(z; 0,\sigma^2),
\]
where $\bone_{S_m}$ is the function having
$\bone_{S_m}(z)=1$ when $z=m(x,\theta)$ for each $\theta\in S_m$, and zero otherwise.
\begin{figure}
	\includegraphics[width=1\textwidth]{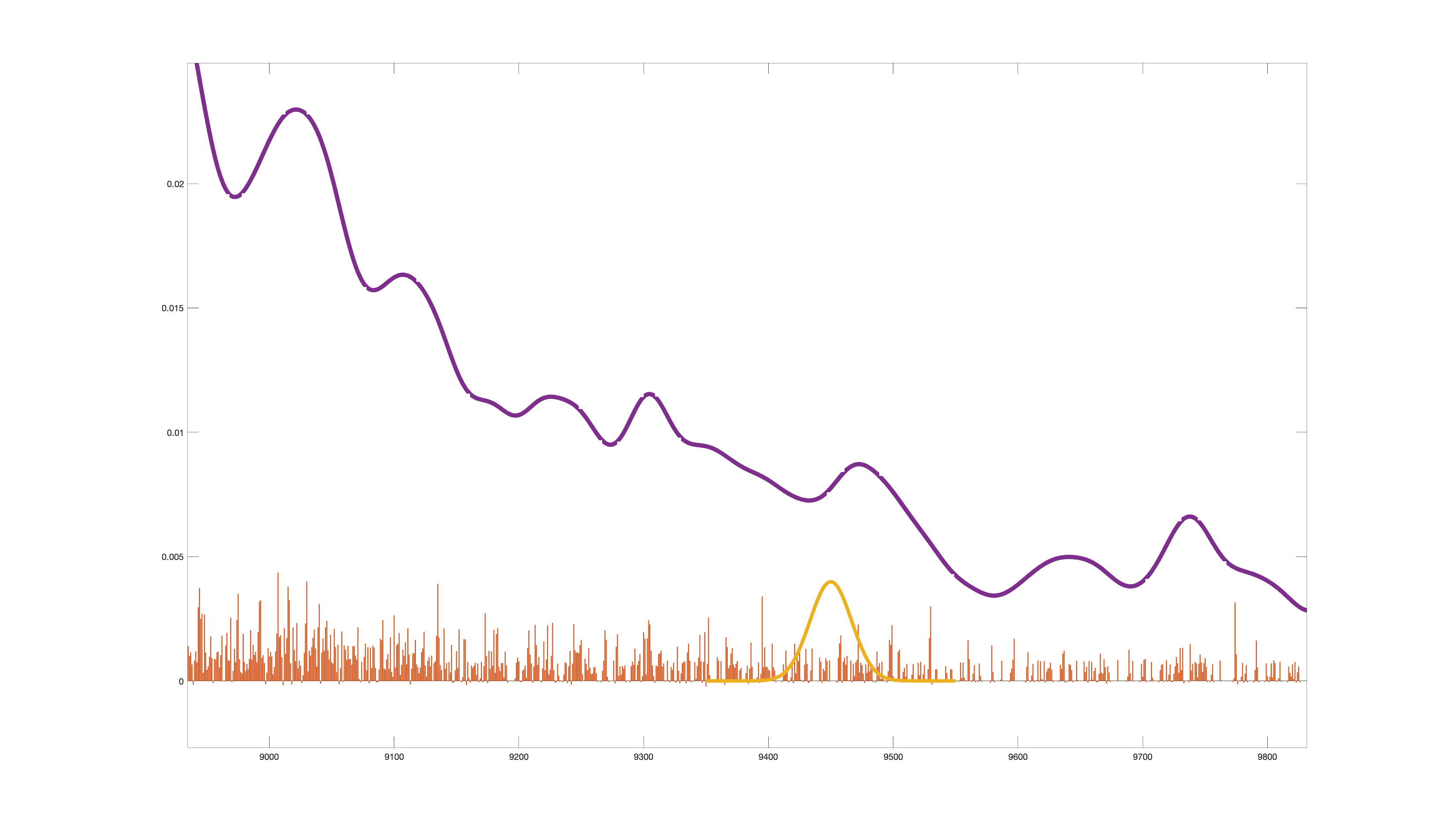}
	\caption{Computing $p(y|m,x)$ via convolution.}
	\label{fig conv}
\end{figure}
By representing $\bone_{S_m}$ and the Gaussian mask $\phi(z; 0,\sigma^2)$
on a fine one-dimensional grid, and using fast convolution,
we can obtain an estimate of $p(y|m,x)$ more quickly
than via numerical integration. We can do better for accuracy than rounding
$m(x,\theta)$ to the nearest grid value by distributing the weight
for each such value across multiple grid points. This still allows fast convolution,
but has the effect of interpolation. In our implementation we distribute weight to the grid values
such that we have cubic interpolation of the Gaussian mask.
The weights, convolution, and mask are shown in Figure~\ref{fig conv}.

Such accuracy might not seem necessary, but
we use {\tt SQP} within {\tt Matlab} to optimize our estimate $\hat H(y|x)$ of $H(y|x)$
with respect to $x$, subject to box constraints on $x$.
This works best if $\hat H(y|x)$ is a smooth function of $x$,
and if a relatively high-accuracy estimate of its gradient $\nabla_x \hat H(y|x)$ is provided.
Some discussion of the gradient computation is given below.

\paragraph{Obtaining $S_m$.}
To obtain a sample $S_m$ of $\theta\sim p(\theta_m)$,
we use Hamiltonian Monte Carlo (HMC), as provided by {\tt Matlab}.
We change the sample only when $p(\theta_m)$ changes.
HMC is relatively fast, but requires $p(\theta)$
to have unbounded support, and requires both $\log p(\theta)$
and $\nabla_\theta \log p(\theta)$ to be provided.
Some discussion of the gradient computation is given below.

The requirement of unbounded support implies that
the easiest approach is to allow e.g. parameters
to be negative even when we know that $\theta_{\rm true}$
has non-negative coordinates. This causes problems
when some parameter is an exponent of an input
coordinate close to zero: the response become unnaturally
large. We ameliorate this by including a term in the prior
for $\theta$ that makes it unlikely that $m(x_0, \theta)$
is extremely large, for some given input point $x_0$.
More generally, we could use changes of variables, or 
rejection methods, to enforce constraints on the parameters
and still use HMC.

\paragraph{Obtaining gradients.}
For HMC, we need $\nabla_{\theta_m} \log p(\theta_m)$, and for
{\tt SQP}, it is helpful to have $\nabla_x H(y|x)$.
(That is, the gradients of our approximations to these functions.)
The former is straightforward to derive and compute, and
the latter is computed via either numerical integration or fast convolution,
just as $H(y|x)$ is. Via the chain rule, for $\nabla_{\theta_m} \log p(\theta_m)$ we need
$\nabla_{\theta} m(x,\theta)$, and for $\nabla_x H(y|x)$,
we need $\nabla_x m(x,\theta)$. In our implementations,
we describe the functional forms as symbolic
expressions, and use {\tt Matlab}'s symbolic toolbox to obtain
functions for computing the models and their gradients.
This is purely as a convenience, as we could write these
functions manually, but this approach allows some scalability
in implementation, and reduces errors.

\section{Numerical study}

\begin{figure}
	\subfigure[Noise $\sigma^2=0.01$]
	{
		\includegraphics[width=1\textwidth]{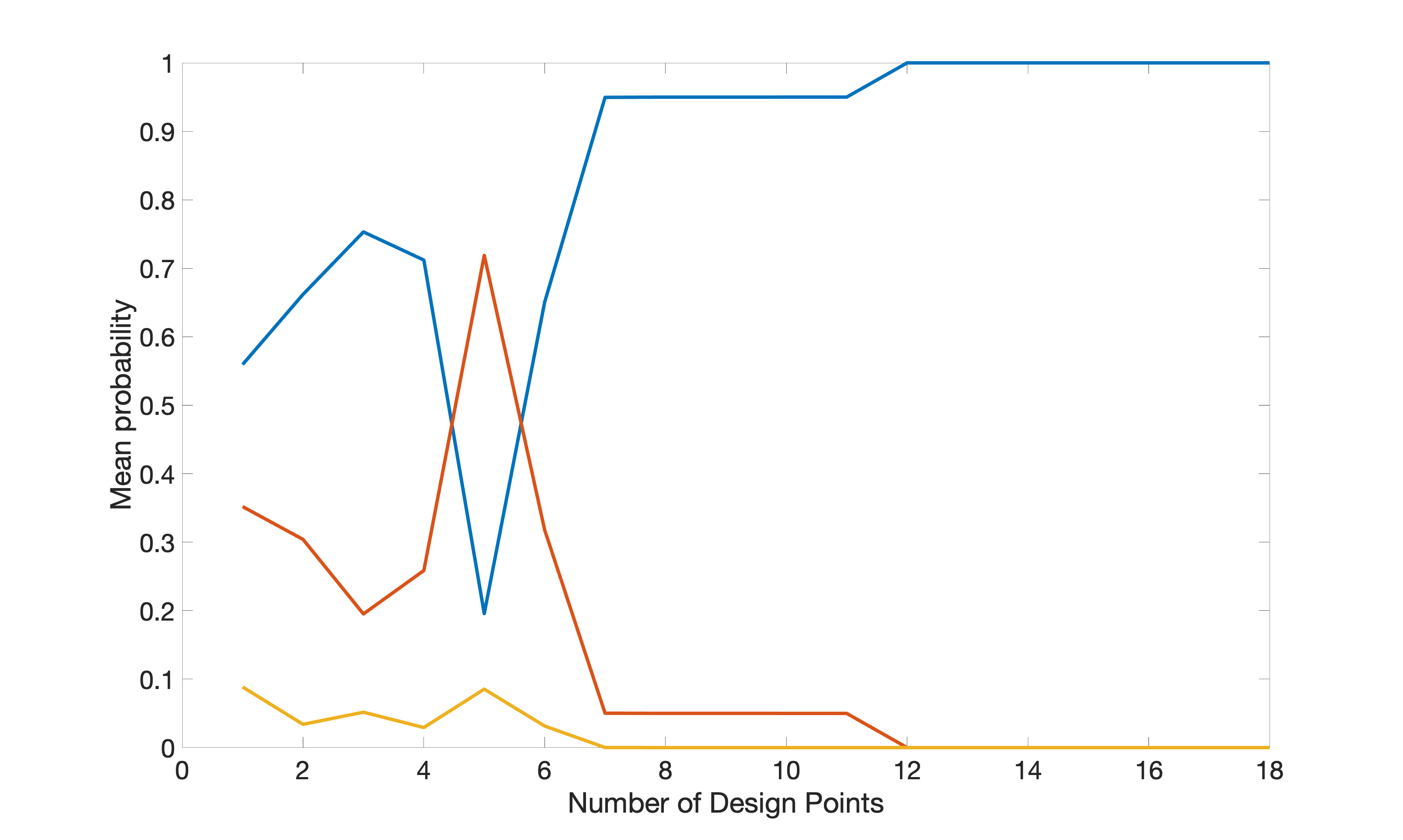}} \\
	\subfigure[Noise $\sigma^2=1$]
	{
		\includegraphics[width=1\textwidth]{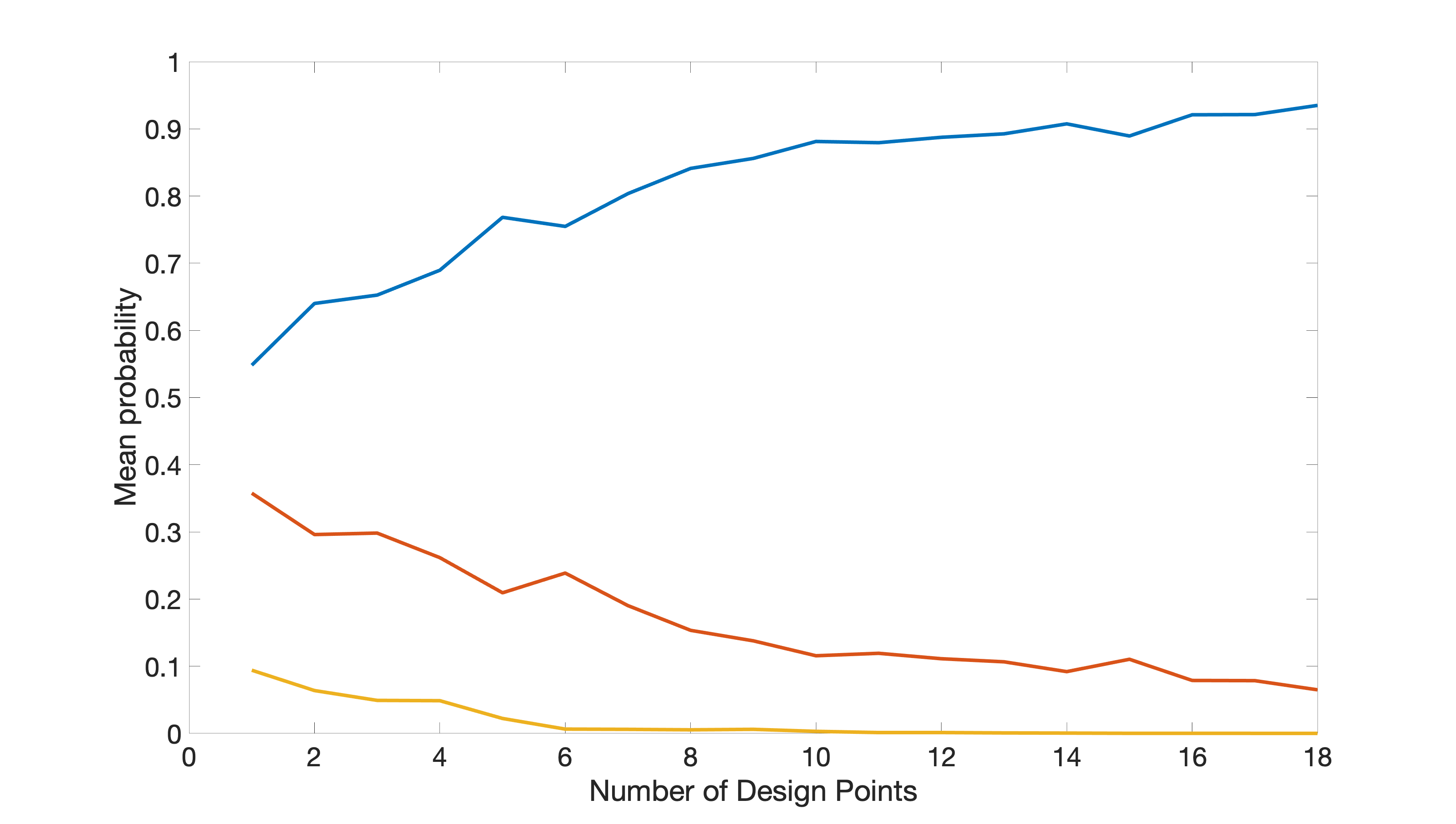}}  
	\caption{Means of model probabilities, mean of twenty trials, over eighteen design points. }
	\label{fig exp}
\end{figure}

We tested and refined our implementation on a challenging-enough small example,
equation~(I.24.6) from Feynman's lecture notes, which is 
\begin{align}
E = cm^{e_1}(\omega^{e_2}+\omega_0^{e_3})z^{e_4}, \label{eq:true_model} 
\end{align}
where $c=1/4$, $e_1=1$ and $e_2=e_3=e_4=2$. 
This model has four inputs $x\defeq (m,\omega,\omega_0,z)$ and five parameters
$\theta\defeq (c,e_1,e_2,e_3,e_4)$. We use three candidate models, the first of which
has the same functional form as the
ground-truth model $\mtrue$ of~\eqref{eq:true_model}.
The other two models are
\begin{align}
E &= cm^{e_1}\omega^{e_2}\omega_0^{e_3}z^{e_4},\\
E &= cm^{e_1}(\omega^{e_2}+z^{e_4})\omega_0^{e_3}. 
\end{align}
We encode the initial values of the parameters of each model, say $\theta_m$ for $m$, in the prior distribution
$p(\theta_m) \sim \calN(\mu_m, \Sigma_m^2)$, where vectors $\mu_m$ have coordinates of magnitude at most 2,
and the matrices $\Sigma_m=I$. We generate 4,000 samples via HMC. The results of two computational experiments
are shown in Figure~\ref{fig exp}. Here the correct model has probability one after twelve trials, for small
noise, and does not quite get to one, for large noise.

\begin{figure}
	\subfigure[Noise $\sigma^2=0.01$]
	{
		\includegraphics[width=0.9\textwidth]{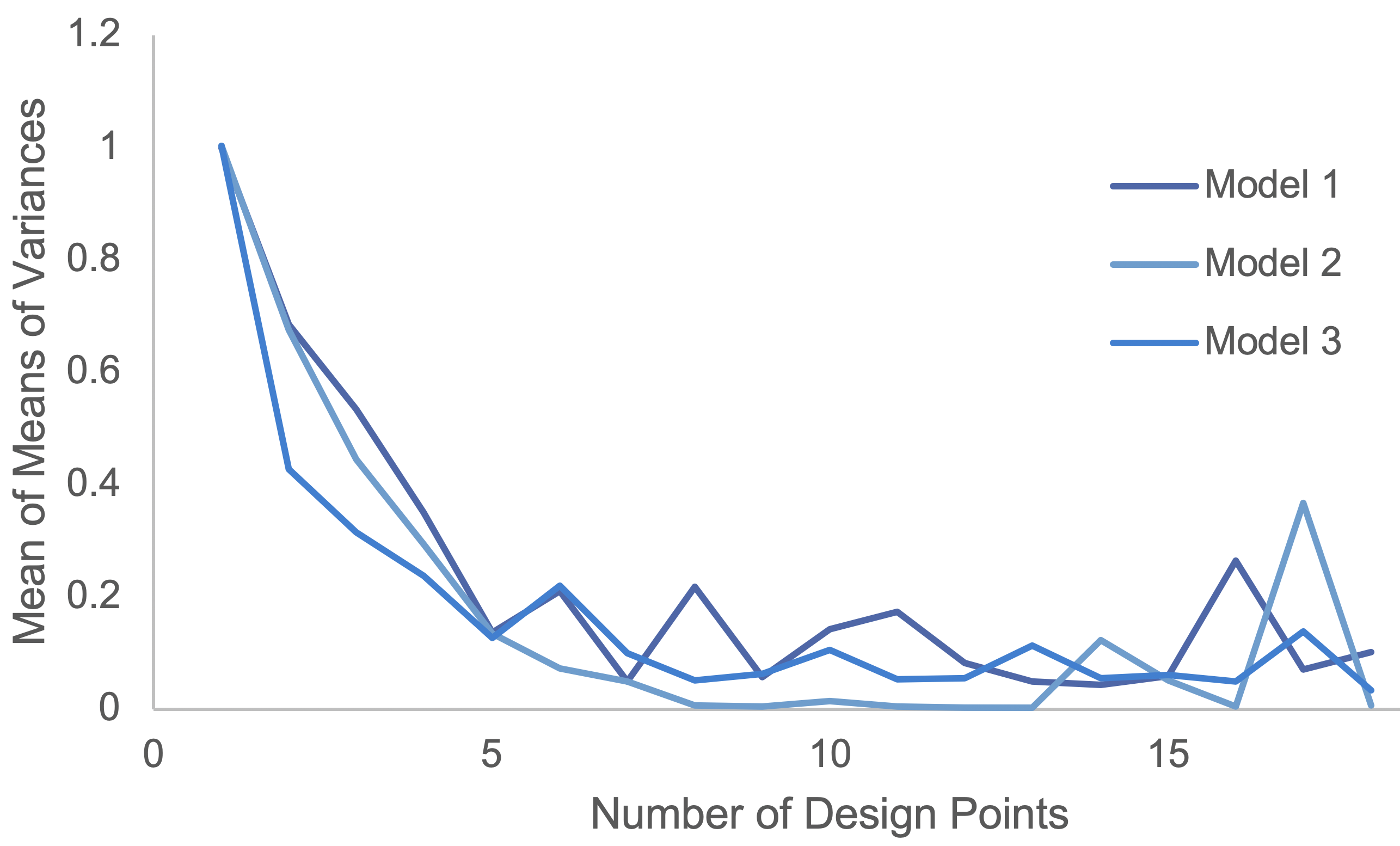}}\\
	\subfigure[Noise $\sigma^2=1$]
	{
		\includegraphics[width=1\textwidth]{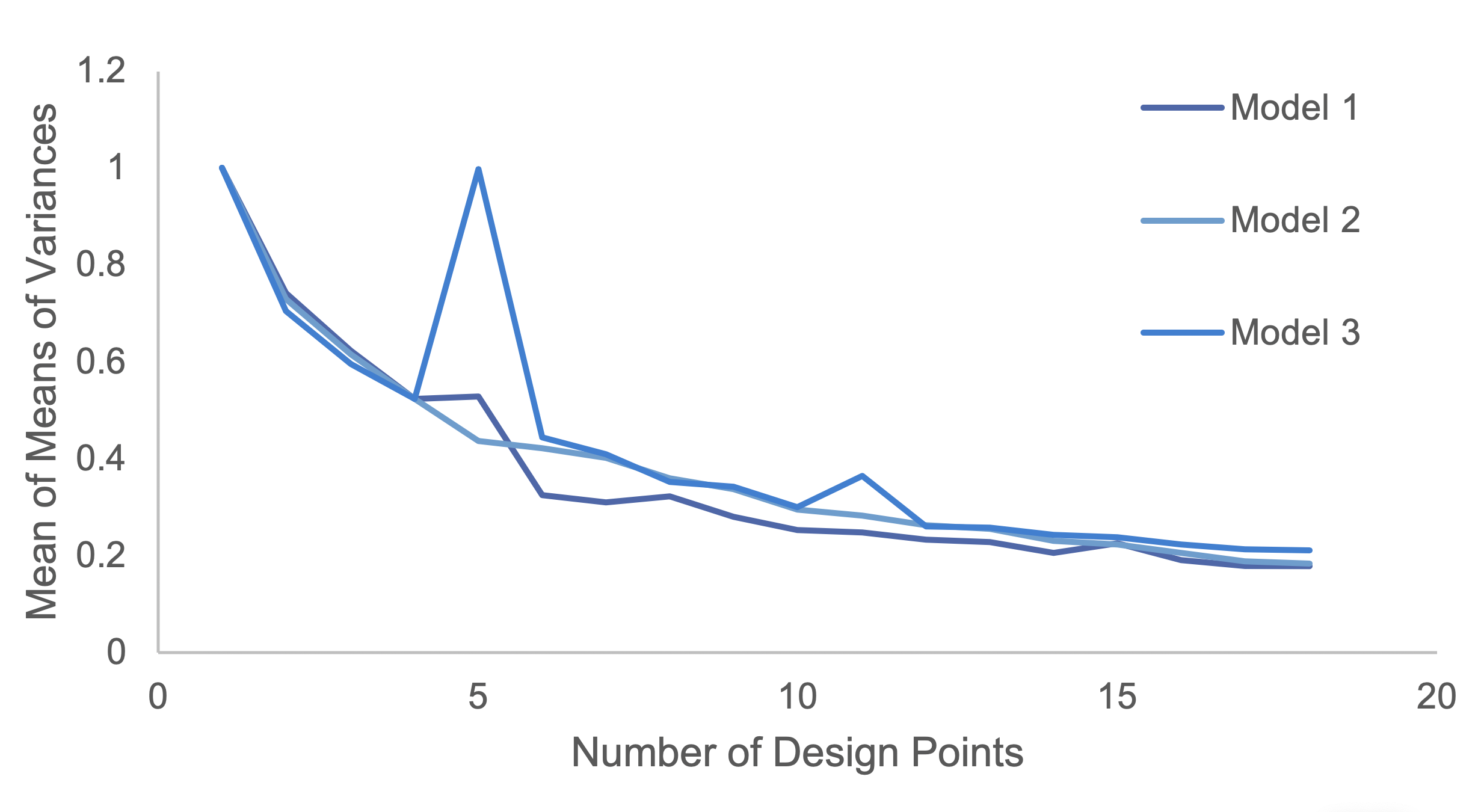}}  
	\caption{Means of per-model parameter variances, mean of twenty trials, over eighteen design points. }
	\label{fig exp var}
\end{figure}

We also tracked the increased knowledge of the parameters, by way of the mean variance of each (the trace of
of the covariance, divided by the number of parameters, as shown in Figure~\ref{fig exp var}

\section{Other choices, further work}\label{subsec B further}

We need to experiment our new formulation  with more settings, and larger problems, although our current runtimes on modest workstations (laptops) are not unwieldy. There are a number of proposed techniques to accelerate Bayesian OED; some of them seem promising for our setting.

\begin{itemize}
    \item Importance sampling of log posteriors (a.k.a. particle methods). As discussed
    by \cite{drovandi2014sequential} for example, these methods reduce the number of times
    Monte Carlo sampling methods are needed, by weighting existing samples to reflect the current
    posterior distribution. So far HMC seems fast enough, but we need to scale up more to see
    if it becomes a bottleneck.
    \item Laplace approximations to the posterior. These use quadratic approximations to the
    posterior log likelihood, in the neighborhood of its maximum. It seems unlikely that such an
    approximation would work well in our case, but this is to be determined.
    \item Approximate Bayesian methods \cite{haber2009numerical,ryan2016review}. These are appropriate in settings where the model
    evaluation itself is expensive; this is not the case here.
    \item Formulations where the design point itself is a random variable, and
    and optimum point is the output of an MCMC process \cite{CMPK10}. We may 
    explore this.
\end{itemize}


\paragraph{Patent Application.} Aspects of this work are covered by U.S. Patent Application \href{https://patents.justia.com/patent/20210334432}{P201809327} ``Experimental Design For Symbolic Model Discovery,'' filed April 2020.

\paragraph{Acknowledgement.}
This work was supported in part by the Defense Advanced Research Projects Agency (DARPA) (PA-18-02-02). The U.S. Government is
authorized to reproduce and distribute reprints for Governmental purposes notwithstanding any copyright notation
thereon.

\newpage

\ifarxiv
\bibliographystyle{alpha}
\fi

\bibliography{OED}
\end{document}